\documentclass{article}
\usepackage[preprint]{neurips_2026}

\usepackage[utf8]{inputenc}
\usepackage[T1]{fontenc}
\usepackage{hyperref}
\usepackage{url}
\usepackage{booktabs}
\usepackage{amsmath}
\usepackage{amsfonts}
\usepackage{nicefrac}
\usepackage{microtype}
\usepackage{xcolor}
\usepackage{graphicx}
\usepackage{enumitem}

\newcommand{\best}[1]{\textbf{#1}}
\title{UniRGBA: Unifying RGB and Alpha in Pretrained Video Diffusion Models}

\author{Lei Xu\textsuperscript{*}, Zhuo Li\textsuperscript{*\ensuremath{\dagger}}, Yongyi Jin\textsuperscript{*}, Aoyu Wang\\
ByteDance\\[2pt]
{\small\ttfamily leixu0325@163.com, lizhuo.726@bytedance.com}\\
{\small\ttfamily jinyongyi@bytedance.com, wangaoyu.0302@bytedance.com}\\[2pt]
{\small\normalfont \textsuperscript{*}Equal contribution. \quad \textsuperscript{\ensuremath{\dagger}}Corresponding author.}}

\begin{document}

\maketitle

\begin{abstract}
Recent advances in video diffusion models have achieved remarkable photorealistic RGB synthesis, yet generating videos with a native alpha channel remains an open challenge. Existing approaches often rely on matting-style data and non-native RGBA designs, causing them to specialize in real-object scenes and to degrade severely on visual-effect content whose semi-transparency is spatially non-local and non-elliptical.
We present \textbf{UniRGBA}, a unified framework whose key insight is that RGBA generation does not require a dedicated alpha encoder or complex latent-space restructuring---a minimal, targeted extension of the existing RGB pathway is sufficient.
\textbf{UniRGBA-VAE} extends a causal video VAE to 4-channel RGBA via zero-initialized channel expansion, symmetric selective layer unfreezing ($\sim$5\% trainable), and a six-term progressive reconstruction loss anchored by a KL reference loss that prevents latent drift, followed by a two-stage schedule that introduces a lightweight GAN objective. We then apply \textbf{standard DoRA} to the pretrained DiT, reusing the unified 4-channel latent without any alpha-specific architectural changes.
On a visual-effect RGBA benchmark, UniRGBA substantially outperforms Wan-Alpha on RGBA reconstruction while preserving RGB fidelity. For generation, UniRGBA outperforms Wan-Alpha on RGBA text-to-video tasks and matches or surpasses the original pretrained Wan2.1 backbone on RGB-only metrics, demonstrating that native alpha support can be integrated without compromising the underlying RGB prior or the model's architectural versatility.
\end{abstract}

\section{Introduction}
\label{sec:intro}

Video generation has witnessed remarkable progress with the advent of large-scale diffusion-based models~\cite{ho2020denoising,rombach2022high,ho2022video,blattmann2023stable,singer2022make,guo2023animatediff,wan2025wan,zheng2024open}. 
Modern video diffusion transformers~\cite{peebles2023scalable} built on powerful visual-language representations~\cite{radford2021learning} have elevated generation quality dramatically. 
State-of-the-art systems such as Wan2.1~\cite{wan2025wan}, CogVideoX~\cite{yang2024cogvideox}, HunyuanVideo~\cite{kong2024hunyuanvideo}, and Seedance~\cite{gao2025seedance} can synthesize photorealistic videos from text or image prompts with stunning temporal coherence. 
However, virtually all existing video generation models operate exclusively in RGB space, producing fully opaque videos without transparency information. 

Alpha channels are indispensable in professional video production workflows. 
Transparent video elements---such as particle effects, fire, smoke, glass reflections, animated stickers, and virtual gifts---must be composited onto arbitrary backgrounds in real time. 
Without native alpha, practitioners must resort to either generating RGB video followed by a separate video matting step~\cite{lin2021real,lin2022robust}, or training specialized RGBA models from scratch, which forfeits the rich visual knowledge embedded in large-scale pretrained RGB models. 
While early efforts like LayerDiffuse~\cite{zhang2024transparent} and Qwen-Image-Layer~\cite{yin2025qwenimagelayered} tackle transparency in still images, recent works such as TransPixeler~\cite{wang2025transpixeler}, Wan-Alpha~\cite{dong2025wan}, AlphaVAE~\cite{wang2025alphavae}, and LayerFlow~\cite{ji2025layerflow} have attempted to extend pretrained models to handle RGBA or layered generation. 
These pioneering works, however, exhibit two critical limitations.

First, \textbf{they rely on matting-style data, leading to failure on visual-effect scenes.} Existing generators are typically trained on datasets where alpha channels are extracted by segmenting solid real-world objects from RGB footage. Consequently, these models implicitly assume the foreground is an opaque, well-defined silhouette. They struggle on \emph{visual-effect} scenes---such as fireworks, particles, flames, and smoke---which dominate real production use cases. In these scenes, semi-transparency is spatially non-local and continuous. Applying matting-biased models to such content frequently causes severe color bleeding and chromatic distortion.

Second, \textbf{they employ non-native RGBA designs that impose heavy overhead and disrupt pretrained RGB priors.} 
TransPixeler~\cite{wang2025transpixeler} appends a separate stream of alpha tokens along the temporal axis, effectively doubling the sequence length processed by the Diffusion Transformer (DiT). 
Wan-Alpha~\cite{dong2025wan} attempts to learn shiftable RGB-A distributions via a bidirectional diffusion loss that restructures the Variational Autoencoder (VAE) latent space. It further injects a Gaussian ellipse mask into diffusion noise sampling for transparency guidance. This design enforces a strong object-centric prior---assuming transparency is roughly elliptical and localized---which fundamentally conflicts with the full-frame nature of visual effects. More importantly, restructuring the latent space misaligns it with the underlying pretrained RGB generator, causing catastrophic forgetting of the original RGB priors and making it difficult to integrate with existing RGB workflows.

Motivated by these limitations, we propose \textbf{UniRGBA}, a unified framework that supports RGBA \emph{natively}---treating alpha simply as a fourth channel of the existing latent representation---through a \emph{minimal, targeted extension} of the pretrained RGB generator. Our goal is to enable high-quality generation of complex visual-effect scenes without compromising the strong RGB fidelity the base model already possesses. Under this paradigm, our architectural intervention is concentrated almost entirely on the VAE, allowing the DiT backbone to remain structurally unchanged.

Specifically, we introduce \textbf{UniRGBA-VAE}, which directly extends the Wan2.1 3D causal VAE to 4-channel RGBA. Instead of shifting the latent distribution or adding auxiliary encoding pathways, we zero-initialize the newly added channel weights in the first and last convolutional layers. This provides a lossless warm start from the pretrained RGB VAE. By employing a symmetric layer unfreezing strategy, we fine-tune only the outermost $2+2$ encoder/decoder layers ($\sim$5\% of parameters), allowing the model to focus on alpha reconstruction while preserving deep RGB semantic representations.

To ensure the newly learned representations do not drift from the original RGB priors, we propose a \textbf{KL-Anchored Progressive Training} strategy. We introduce a KL reference loss ($\mathcal{L}_{\text{kl\_ref}}$) that softly anchors the fine-tuned posterior to the frozen original RGB posterior. The VAE is trained using a six-term progressive reconstruction objective under a two-stage schedule. The first stage focuses on stabilizing the joint RGB-A manifold, while the second stage adds a lightweight GAN loss to the composited outputs to sharpen fine-grained details typical of visual effects (e.g., thin smoke wisps and particle edges).

Because UniRGBA-VAE preserves the structure and distribution of the original latent space, we can perform \textbf{Standard DoRA DiT Adaptation} directly on the unified 4-channel latents. We insert standard, off-the-shelf DoRA modules into the pretrained Wan2.1 DiT without any alpha-specific architectural changes (e.g., no extra alpha tokens or shifted noise schedules). This design supports RGBA text-to-video generation directly and also admits an image-to-video (I2V) extension by injecting reference-image conditions in the same unified latent space; due to the lack of standardized RGBA I2V benchmarks and reproducible baselines, we leave I2V evaluation to qualitative supplementary visualizations.

Extensive evaluations on visual-effect RGBA data demonstrate the effectiveness of this minimal intervention. UniRGBA-VAE substantially outperforms Wan-Alpha on RGBA reconstruction while strictly preserving RGB fidelity. For generation, our full pipeline outperforms Wan-Alpha on RGBA text-to-video tasks, and on RGB-only generation, it achieves quality on par with---and on several metrics surpassing---the original pretrained Wan2.1 backbone. This confirms that native alpha support can be achieved efficiently without sacrificing pretrained RGB capabilities.

Our main contributions are summarized as follows:
\begin{itemize}
\item We propose \textbf{UniRGBA}, a unified and minimal framework for native RGBA video generation. By treating alpha as a fourth channel and employing a KL-anchored progressive training strategy on the VAE, we preserve the pretrained RGB latent manifold to a large extent. This alignment enables RGBA Text-to-Video (T2V) generation using standard, off-the-shelf DoRA DiT adapters without requiring bespoke alpha architectures.
\item \textbf{State-of-the-Art Reconstruction Performance:} UniRGBA-VAE establishes a new state-of-the-art for RGBA video reconstruction on challenging visual-effect content. It substantially outperforms existing baselines like Wan-Alpha across both RGB and alpha metrics, delivering highly accurate transparency details without suffering from the severe color bleeding and chromatic drift caused by latent distribution shifting.
\item \textbf{State-of-the-Art Generation Performance:} Extensive evaluations demonstrate that our full pipeline significantly outperforms current RGBA generation baselines like Wan-Alpha on T2V tasks. More importantly, unlike prior methods that suffer from catastrophic forgetting, UniRGBA strictly preserves the backbone's original capabilities, achieving RGB-only generation quality on par with---and on several metrics surpassing---the original pretrained Wan2.1 model.
\end{itemize}

\section{Related Work}
\label{sec:related}

\paragraph{Large-Scale Video Generation.}
The landscape of video generation has been revolutionized by diffusion models~\cite{ho2020denoising,rombach2022high,ho2022video} and the emergence of Diffusion Transformers (DiT)~\cite{peebles2023scalable}. Early systems like Make-A-Video~\cite{singer2022make}, VideoComposer~\cite{wang2023videocomposer}, and Stable Video Diffusion~\cite{blattmann2023stable} extended spatial U-Nets with temporal modules. More recently, large-scale DiT-based architectures trained on massive video-text datasets have set new benchmarks for visual fidelity, motion dynamics, and prompt adherence. Prominent examples include Sora~\cite{brooks2024video}, Open-Sora~\cite{zheng2024open}, CogVideoX~\cite{yang2024cogvideox}, HunyuanVideo~\cite{kong2024hunyuanvideo}, LTX-Video~\cite{HaCohen2024LTXVideo}, Mochi~\cite{genmo2024mochi}, Vidu~\cite{bao2024viduhighlyconsistentdynamic}, Seedance~\cite{gao2025seedance}, and Wan2.1~\cite{wan2025wan}. Despite their unprecedented capabilities in modeling complex physics and semantics, these foundation models operate exclusively in the RGB color space. Generating videos with an intrinsic alpha channel for transparency---a critical requirement for downstream compositing in professional video production---remains largely unsupported by these general-purpose pipelines.

\paragraph{Transparent Image and Video Generation.}
Generating transparency has traditionally been treated as a downstream matting or segmentation task~\cite{lin2021real,lin2022robust,ravi2024sam2}, rather than a native generative process. In the image domain, LayerDiffuse~\cite{zhang2024transparent} pioneered transparent image generation by fine-tuning latent diffusion models to output RGBA, while other works leverage ControlNet~\cite{zhang2023adding} to guide foreground generation, and recent efforts like AlphaVAE~\cite{wang2025alphavae} and Qwen-Image-Layer~\cite{yin2025qwenimagelayered} have further explored end-to-end unified layer representations.
Extending this to the temporal domain is highly challenging. LayerFlow~\cite{ji2025layerflow} proposes a flow-based approach to generate independent transparent foregrounds and backgrounds. TransPixeler~\cite{wang2025transpixeler} augments pretrained video generators with alpha-aware tokens and adapter-based fine-tuning. Wan-Alpha~\cite{dong2025wan} introduces shifted RGB-A distributions to support transparent video generation, utilizing a bidirectional diffusion loss and a Gaussian ellipse mask for noise guidance. 
However, both approaches exhibit notable drawbacks. First, they are typically trained on datasets constructed via video matting pipelines~\cite{lin2021real}, which inherently bias the models toward solid, real-world objects (e.g., humans or pets). As a result, they struggle significantly with visual-effect scenes (e.g., fireworks, flames, and smoke) where transparency is spatially continuous and non-elliptical. Second, these architectures do not treat RGBA as a fully native modality: appending temporal alpha tokens (TransPixeler) increases computational overhead, while restructuring the latent space (Wan-Alpha) disrupts the underlying RGB generator's prior. In contrast, UniRGBA treats the alpha channel as a native fourth dimension within the original latent space, preserving the pretrained RGB manifold while extending its capability to complex visual effects.

\paragraph{Variational Autoencoders for Video Latent Modeling.}
Variational autoencoders~\cite{kingma2013auto} serve as the essential compression backbone for latent diffusion systems~\cite{rombach2022high,podell2023sdxl}. In the video domain, the VAE must effectively compress spatio-temporal signals while preserving temporal coherence and high-frequency details. Recent efforts such as MAGVIT~\cite{yu2023magvit}, MAGVIT-v2~\cite{yu2023language}, CV-VAE~\cite{zhao2024cv}, and OmniTokenizer~\cite{wang2024omnitokenizer} have explored advanced 3D tokenization and causal architectures to improve video compression limits. The Wan-VAE~\cite{wan2025wan} adopts a highly efficient 3D causal structure that provides exceptional RGB reconstruction, making it an ideal backbone. 
While prior works focus almost entirely on RGB compression, adapting a pretrained VAE to a new modality (RGBA) poses a severe risk of catastrophic forgetting, which can render the accompanying pretrained DiT obsolete. UniRGBA-VAE addresses this by extending the pretrained Wan-VAE via zero-initialized channel expansion, symmetric selective layer unfreezing, and a KL reference loss. This minimal-intervention approach successfully models complex alpha structures without destroying the latent distribution expected by the pretrained video generator.

\section{Method}
\label{sec:method}

\begin{figure}[t]
    \centering
    \includegraphics[width=\textwidth]{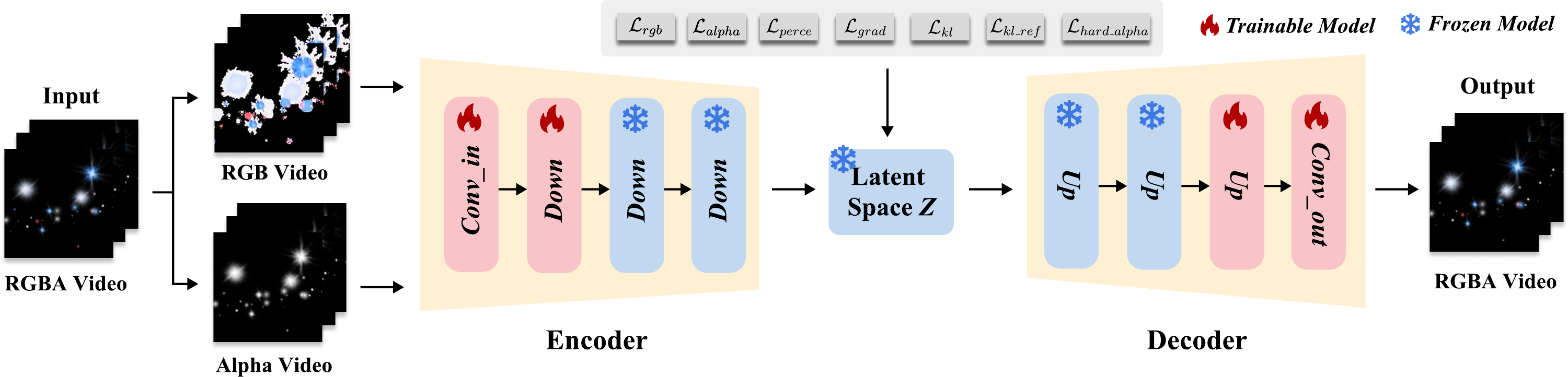}
    \caption{\textbf{Overview of UniRGBA-VAE.} UniRGBA-VAE extends the Wan2.1 VAE to 4-channel RGBA through zero-initialized channel expansion at Conv\_in/Conv\_out with selective layer fine-tuning (orange = trainable, blue = frozen). Training losses include pixel-level (L\textsubscript{rgb}, L\textsubscript{alpha}, L\textsubscript{comp}, L\textsubscript{hard\_a}), perceptual (L\textsubscript{perce}), and regularization (L\textsubscript{kl}, L\textsubscript{kl\_ref}) terms.}
    \label{fig:vae_architecture}
\end{figure}

\begin{figure}[t]
    \centering
    \includegraphics[width=\textwidth]{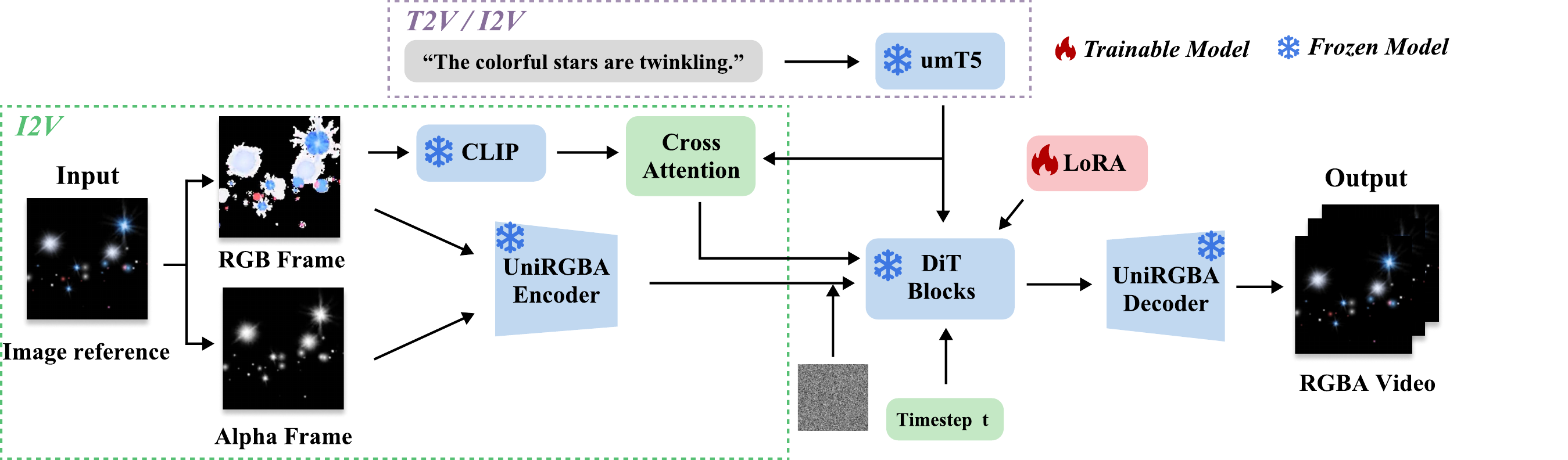}
    \caption{\textbf{Overview of Alpha-aware downstream adaptation.} For T2V, frozen umT5 text features and timestep embeddings condition adapter-tuned DiT blocks, whose denoised latents are decoded by UniRGBA-VAE. For I2V, a reference RGBA frame is additionally encoded into visual conditions in the same latent space and injected into the shared DiT backbone.}
    \label{fig:dit_architecture}
\end{figure}

Figures~\ref{fig:vae_architecture} and~\ref{fig:dit_architecture} provide an overview of our framework. We describe each component below.

\paragraph{Framework Overview.}
UniRGBA is a two-component framework that upgrades a pretrained RGB video generator into a native RGBA video generator through a \emph{minimal, targeted} extension of the existing RGB pathway, without introducing any alpha-specific architectural module or restructuring the latent space.
At the \emph{representation} level, \textbf{UniRGBA-VAE} (Section~\ref{sec:vae}) extends the Wan2.1 3D causal VAE to native 4-channel RGBA by zero-initializing the added channel in the first and last convolutional layers, giving a lossless warm start from the pretrained RGB VAE; only the outermost $2+2$ encoder/decoder layers ($\sim$5\% of parameters) are fine-tuned, and the posterior is softly anchored to the frozen original RGB posterior by a KL reference loss $\mathcal{L}_{\text{kl\_ref}}$ (Section~\ref{sec:training_objective}) that prevents latent manifold drift during adaptation. Training proceeds on a six-term progressive reconstruction objective (Eq.~\ref{eq:total}) under a two-stage cosine schedule: Stage-1 is reconstruction-only, and Stage-2 additionally introduces a 0.02-weighted lightweight GAN objective (Eq.~\ref{eq:stage1}--\ref{eq:stage2}) to refine fine-grained RGB and alpha details.
At the \emph{generation} level, we then apply \textbf{standard DoRA adaptation} of the pretrained Wan2.1 DiT (Section~\ref{sec:dit}) directly on the unified 4-channel latent produced by UniRGBA-VAE, introducing no alpha-specific tokens, conditioning pathways, or latent-distribution restructuring.
Together, these two components form a unified pipeline in which RGBA support is added almost entirely on the VAE side, while the DiT backbone is reused as-is and adapted with off-the-shelf parameter-efficient fine-tuning.

\subsection{UniRGBA-VAE Architecture}
\label{sec:vae}

\paragraph{Channel Expansion with Zero Initialization.}
We build upon the pretrained Wan2.1 3D causal VAE, which maps RGB video $\mathbf{x} \in \mathbb{R}^{T \times H \times W \times 3}$ to a latent representation $\mathbf{z} \in \mathbb{R}^{T/4 \times H/8 \times W/8 \times 16}$.
To support RGBA input $\mathbf{x} \in \mathbb{R}^{T \times H \times W \times 4}$, we expand the first convolutional layer (Conv\_in) from 3 to 4 input channels and the final layer (Conv\_out) from 3 to 4 output channels.
Critically, the weights for the new alpha channel are \emph{zero-initialized}, which provides a lossless warm start: when the alpha channel is set to 1 (fully opaque), the model output is \emph{identical} to the original Wan-VAE at initialization.
This initialization strategy is in contrast to distribution-shifting approaches~\cite{dong2025wan}, which require additional bidirectional diffusion losses to restructure the latent space; our approach requires no such restructuring, as the RGB latent manifold is preserved by construction.

\paragraph{Symmetric Selective Layer Unfreezing.}
Fine-tuning the entire VAE risks catastrophic forgetting of the strong RGB reconstruction capability acquired during large-scale pretraining.
To adapt the pretrained VAE to 4-channel inputs without destroying this prior, we propose a symmetric unfreezing strategy.
We train only the first $k$ encoder layers and the last $k$ decoder layers while keeping all intermediate layers frozen.
Specifically, we unfreeze Conv\_in + DownBlock1 (encoder side) and UpBlock3 + Conv\_out (decoder side), corresponding to $k=2$.
By symmetrically unfreezing both the input encoder layers and the output decoder layers, the model can jointly optimize spatial-transparency feature extraction and high-frequency boundary reconstruction, leading to more coherent overall RGBA representations without compromising the deeply embedded RGB priors.
Under this symmetric $2+2$ configuration, only $\sim$5\% of the total parameters are trainable.

\subsection{KL-Anchored Progressive Training}
\label{sec:training_objective}

\paragraph{Progressive Multi-Loss Objective.}
We design a comprehensive loss function with six complementary terms, each targeting a specific aspect of RGBA reconstruction quality:

\vspace{-0.5em}
\begin{align}
\mathcal{L}_{\text{rgb}} &= \| (\hat{\mathbf{x}}_{\text{rgb}} \odot \mathbf{a}) - (\mathbf{x}_{\text{rgb}} \odot \mathbf{a}) \|_1 \label{eq:lrgb} \\
\mathcal{L}_{\text{alpha}} &= \| \hat{\mathbf{a}} - \mathbf{a} \|_1 \label{eq:lalpha} \\
\mathcal{L}_{\text{comp}} &= \| (\hat{\mathbf{x}}_{\text{rgb}} \odot \hat{\mathbf{a}} + \mathbf{b} \odot (1 - \hat{\mathbf{a}})) - (\mathbf{x}_{\text{rgb}} \odot \mathbf{a} + \mathbf{b} \odot (1 - \mathbf{a})) \|_1 \label{eq:lcomp} \\
\mathcal{L}_{\text{perce}} &= \text{LPIPS}(\hat{\mathbf{c}}_{\text{comp}}, \mathbf{c}_{\text{comp}}) \label{eq:lperce} \\
\mathcal{L}_{\text{kl}} &= D_{\text{KL}}(q(\mathbf{z}|\mathbf{x}) \| \mathcal{N}(\mathbf{0}, \mathbf{I})) \label{eq:lkl} \\
\mathcal{L}_{\text{kl\_ref}} &= D_{\text{KL}}(q(\mathbf{z}|\mathbf{x}) \| q_{\text{ref}}(\mathbf{z}|\mathbf{x}_{\text{rgb}})) \label{eq:lklref}
\end{align}
\vspace{-0.5em}

where $\hat{\mathbf{x}}_{\text{rgb}}$ and $\hat{\mathbf{a}}$ denote predicted RGB and alpha channels, $\mathbf{b}$ is a randomly sampled background color used for compositing supervision, $q_{\text{ref}}$ is the posterior of the \emph{frozen} original Wan-VAE, and $\mathcal{L}_{\text{perce}}$ uses LPIPS~\cite{zhang2018unreasonable} as the perceptual distance.
For hard-alpha regions (where $\mathbf{a} \in \{0, 1\}$), we additionally apply:
\begin{equation}
\mathcal{L}_{\text{hard\_alpha}} = \| (\hat{\mathbf{a}} \odot \mathbf{m}_{\text{hard}}) - (\mathbf{a} \odot \mathbf{m}_{\text{hard}}) \|_1
\label{eq:lhard}
\end{equation}
where $\mathbf{m}_{\text{hard}}$ is a binary mask identifying regions with alpha values close to 0 or 1. The progressive training objective combines six pixel-, perceptual-, and regularization-level terms with the hard-alpha auxiliary:
\begin{equation}
\mathcal{L}_{\text{recon}} = \mathcal{L}_{\text{rgb}} + \mathcal{L}_{\text{alpha}} + 0.5 \cdot \mathcal{L}_{\text{comp}} + 0.5 \cdot \mathcal{L}_{\text{perce}} + 10^{-6} \cdot \mathcal{L}_{\text{kl}} + 10^{-10} \cdot \mathcal{L}_{\text{kl\_ref}} + 0.5 \cdot \mathcal{L}_{\text{hard\_alpha}}
\label{eq:total}
\end{equation}

\paragraph{Latent Distribution Anchoring via \texorpdfstring{$\mathcal{L}_{\text{kl\_ref}}$}{L\_kl\_ref}.}

A critical challenge in fine-tuning large-scale VAEs is latent manifold drift: as the model adapts to 4-channel input, the learned posterior $q(\mathbf{z}|\mathbf{x})$ may diverge from the pretrained RGB posterior, causing the DiT (which was trained on the original latent space) to receive out-of-distribution inputs at inference time.
We introduce $\mathcal{L}_{\text{kl\_ref}}$, a KL divergence from the fine-tuned posterior to the frozen reference VAE posterior $q_{\text{ref}}$.
Despite its extremely small weight ($10^{-10}$), this term acts as a \emph{soft anchor} to the original RGB latent distribution: the newly learned alpha representations are constrained to be "parasitic" additions to, rather than destructive modifications of, the pretrained RGB latent space.
This minimal soft anchor provides sufficient regularization to prevent catastrophic forgetting and ensures that latent space stability is maintained throughout the fine-tuning process.

\paragraph{Two-Stage GAN-Augmented Training.}
To further refine fine-grained reconstruction quality (fine edges of particles, smoke wisps, and semi-transparent textures), we train the final reconstruction model in two stages under a single cosine learning-rate schedule.
We introduce a lightweight patch-based adversarial objective $\mathcal{L}_{\text{GAN}}$ on the alpha-composited output only in the second stage.
\emph{Stage-1} trains the VAE from the zero-initialized RGBA warm start using only the reconstruction loss in Eq.~\ref{eq:total}:
\begin{equation}
\mathcal{L}_{\text{stage-1}} = \mathcal{L}_{\text{recon}}.
\label{eq:stage1}
\end{equation}
\emph{Stage-2} continues training from the Stage-1 checkpoint along the same cosine schedule, augmenting the objective with a small-weight adversarial term to refine fine-grained details after the reconstruction loss has stabilized:
\begin{equation}
\mathcal{L}_{\text{stage-2}} = \mathcal{L}_{\text{recon}} + 0.02 \cdot \mathcal{L}_{\text{GAN}}.
\label{eq:stage2}
\end{equation}
Adding the GAN objective on top of an already-strong reconstruction-only Stage-1 checkpoint substantially improves pixel-level RGB/alpha fidelity and perceptual quality, effectively capturing the high-frequency structural details essential for visual-effect scenes.

\subsection{Alpha-Aware DiT Adaptation for T2V and I2V}
\label{sec:dit}

To generate RGBA videos, we adapt the pretrained Wan2.1 diffusion transformer on top of our expanded UniRGBA-VAE. The goal is to determine whether the unified RGBA latent space can support downstream generation with only lightweight parameter-efficient tuning, while keeping the large text-video backbone largely frozen. We deliberately do not introduce any RGBA-specific architectural modification (no alpha tokens, no extra conditioning pathway, no dedicated alpha attention).

\paragraph{T2V adaptation.}
During T2V training, the text prompt is encoded by the frozen umT5 encoder, UniRGBA-VAE provides RGBA latent targets, and the DiT denoises latent videos conditioned on text embeddings and timestep embeddings. We insert lightweight DoRA-style adapters~\cite{liu2024dora} into the attention projections and feed-forward layers of all DiT blocks, while freezing the remaining backbone weights. Concretely, we adapt the standard Wan2.1 DiT modules \texttt{q}, \texttt{k}, \texttt{v}, \texttt{o}, \texttt{ffn.0}, and \texttt{ffn.2}. This design preserves the pretrained motion prior and text-video alignment behavior while enabling RGBA-specific generation. As demonstrated by Wan-Alpha~\cite{dong2025wan}, DoRA (Weight-Decomposed Low-Rank Adaptation) acts as a stricter regularizer on the pretrained behavior than standard LoRA, making it superior for preserving the original RGB generation capabilities while learning new alpha dynamics. Therefore, we exclusively adopt DoRA for all our downstream adaptations.

\paragraph{I2V extension.}
The same UniRGBA latent interface also supports image-to-video generation without introducing a separate alpha-specific branch. Given a reference RGBA image, we encode it with UniRGBA-VAE to obtain reference latent tokens. Following the conditioning design of Wan2.1 I2V, the same reference image also provides frozen visual features from the image encoder, which are concatenated with the text context. The DiT then predicts a sequence of RGBA latents conditioned on text, timestep, and reference-image cues, after which UniRGBA-VAE decodes the result into an RGBA video. T2V and I2V therefore share the same downstream backbone: T2V uses text-only conditioning, whereas I2V augments this backbone with reference-image conditions in the same unified RGBA latent space.

\section{Experiments}
\label{sec:experiments}

\subsection{Experimental Setup}
\label{sec:setup}

We curate an internal RGBA video dataset from historical special-effect asset packages. The datasets and detailed training schedules (including learning rates, batch sizes, and multi-stage loss configurations) are provided in Appendix~\ref{sec:appendix_setup}.

\subsection{Evaluation Metrics and Baselines}
\label{sec:setup_metrics}

For VAE reconstruction, we report PSNR, SSIM, and LPIPS. We decompose PSNR into: PSNR\textsubscript{rgb}, PSNR\textsubscript{alpha}, PSNR\textsubscript{rgb\_mask}, and PSNR\textsubscript{alpha\_mask}.
For downstream generation, we report \emph{Aesthetic Quality}, \emph{Text Alignment}, \emph{Naturalness}, \emph{Motion Smoothness}, and \emph{Temporal Flickering}. For RGB-only generation, we additionally measure \emph{Dynamic Degree} and \emph{Temporal Spatial Consistency}.
We evaluate RGBA capabilities against Wan-Alpha~\cite{dong2025wan}, and RGB capabilities against the original Wan2.1~\cite{wan2025wan} to assess the preservation of the pretrained RGB prior. 
Detailed descriptions of the datasets, preprocessing, and training schedules are provided in Appendix~\ref{sec:appendix_setup}.

\subsection{Main Results: VAE Reconstruction}
\label{sec:main_results}

Table~\ref{tab:main} summarizes the main reconstruction results. Unless otherwise specified, ablation studies in this section use the 1K RGBA ablation set, while Table~\ref{tab:main} reports the final two-stage reconstruction model trained on the 17K high-quality RGBA subset. Stage-1 already outperforms Wan-Alpha on all overall metrics without adversarial supervision; Stage-2 further delivers large gains on PSNR, PSNR\textsubscript{rgb}, and PSNR\textsubscript{alpha} by adding a lightweight GAN objective. On RGB-only WebVid reconstruction, UniRGBA-VAE demonstrates robust RGB preservation, maintaining highly competitive performance compared to the original Wan2.1 VAE.

\paragraph{Superior RGB and Alpha Fidelity.}
Compared to Wan-Alpha, our Stage-2 UniRGBA-VAE achieves significant improvements across all overall metrics (e.g., PSNR 25.00~$\rightarrow$~29.16, SSIM 0.891~$\rightarrow$~0.932). Crucially, this gain is seen in both RGB and alpha pathways (PSNR\textsubscript{rgb} 24.42~$\rightarrow$~29.75; PSNR\textsubscript{alpha} 30.50~$\rightarrow$~34.45). This demonstrates that our zero-initialized expansion and symmetric unfreezing strategy preserves the pretrained RGB manifold by construction. In contrast, Wan-Alpha employs a bidirectional latent-shifting objective that explicitly forces the latent distribution outward to accommodate alpha, which inherently misaligns with the pretrained RGB decoder's expectations, leading to the severe color bleeding and chromatic drift (manifested as low PSNR\textsubscript{rgb}) observed in their results.

\paragraph{Effect of the Stage-2 GAN-augmented schedule.}
Stage-2 introduces adversarial supervision on the alpha-composited output, yielding substantial gains on global sharpness and pixel-level metrics (PSNR 26.25~$\rightarrow$~29.16, LPIPS 0.0808~$\rightarrow$~0.0801). This comes with a marginal trade-off on local mask-region fidelity (PSNR\textsubscript{rgb\_mask} 22.76~$\rightarrow$~22.49), a common characteristic of GAN objectives favoring globally plausible textures over strict pixel-wise agreement. We report both stages to make this trade-off explicit.

\paragraph{Backward compatibility on pure RGB data.}
On the RGB-only WebVid benchmark, UniRGBA-VAE performs on par with or slightly better than the original Wan2.1 VAE baseline (SSIM 0.945 vs.\ 0.964; LPIPS 0.0401 vs.\ 0.0213). This minor PSNR gap is an acceptable trade-off for enabling RGBA capabilities. By zero-initializing the alpha pathway, our encoder produces latents \emph{identical} to the original Wan-VAE when fed pure RGB video ($\alpha=1$), making UniRGBA-VAE a true drop-in replacement.

\begin{table}[t]
\centering
\caption{\textbf{Main reconstruction results on RGBA benchmark and RGB-only WebVid.} \best{Best} results are shown in bold. The WebVid comparison confirms that the minor PSNR drop versus the original Wan2.1 VAE is an acceptable trade-off for enabling RGBA capabilities.}
\label{tab:main}
\vspace{0.5em}
\resizebox{\textwidth}{!}{
\begin{tabular}{l c c c c c}
\toprule
\textbf{Metric} & \multicolumn{3}{c}{\textbf{Ours RGBA dataset}} & \multicolumn{2}{c}{\textbf{WebVid dataset}} \\
\cmidrule(lr){2-4}\cmidrule(lr){5-6}
& \textbf{Wan-Alpha} & \textbf{UniRGBA-VAE (Stage-1)} & \textbf{UniRGBA-VAE (Stage-2)} & \textbf{UniRGBA-VAE} & \textbf{Wan2.1-VAE} \\
\midrule
\textbf{PSNR}$\uparrow$ & 25.00 & 26.25 & \best{29.16} & 33.41 & \best{36.73} \\
\textbf{SSIM}$\uparrow$ & 0.891 & \best{0.933} & 0.932 & 0.945 & \best{0.964} \\
\textbf{LPIPS}$\downarrow$ & 0.0936 & 0.0808 & \best{0.0801} & 0.0401 & \best{0.0213} \\
\textbf{PSNR\textsubscript{rgb}}$\uparrow$ & 24.42 & 26.16 & \best{29.75} & 33.41 & \best{36.73} \\
\textbf{PSNR\textsubscript{rgb\_mask}}$\uparrow$ & 20.83 & \best{22.76} & 22.49 & -- & -- \\
\textbf{PSNR\textsubscript{alpha}}$\uparrow$ & 30.50 & 32.56 & \best{34.45} & -- & -- \\
\textbf{PSNR\textsubscript{alpha\_mask}}$\uparrow$ & \best{22.54} & 21.91 & 22.02 & -- & -- \\
\bottomrule
\end{tabular}
}
\end{table}
The results support three conclusions. First, even the Stage-1 UniRGBA-VAE recipe, trained purely with our six-term progressive reconstruction loss (no GAN), substantially outperforms Wan-Alpha on all overall metrics. Second, adding a lightweight GAN objective only in Stage-2 yields substantial further gains on PSNR, PSNR\textsubscript{rgb}, PSNR\textsubscript{alpha}, and LPIPS, at the cost of a small trade-off on SSIM and PSNR\textsubscript{rgb\_mask}. Third, on RGB-only WebVid, UniRGBA-VAE demonstrates strong preservation of RGB priors, maintaining competitive performance compared to the original Wan2.1 VAE.

\subsection{Loss Ablation}
\label{sec:ablation_loss}

Table~\ref{tab:loss_ablation} presents our progressive loss ablation, starting from a minimal baseline and incrementally adding loss terms. All experiments use 1K training videos at 832$\times$480 resolution with 5 frames.

\begin{table}[t]
\centering
\caption{\textbf{Progressive loss ablation.} Each row adds one loss component. $\mathcal{L}_{\text{kl\_ref}}$ provides the largest single improvement, and $\mathcal{L}_{\text{comp}}$ is essential for alpha-aware compositing quality.}
\label{tab:loss_ablation}
\vspace{0.5em}
\resizebox{\textwidth}{!}{
\begin{tabular}{cl c c c c c}
\toprule
\textbf{ID} & \textbf{Configuration} & \textbf{PSNR}$\uparrow$ & \textbf{PSNR\textsubscript{rgb}}$\uparrow$ & \textbf{PSNR\textsubscript{alpha}}$\uparrow$ & \textbf{SSIM}$\uparrow$ & \textbf{LPIPS}$\downarrow$ \\
\midrule
E001 & $\mathcal{L}_{\text{rgb}} + \mathcal{L}_{\text{alpha}} + \mathcal{L}_{\text{perce}} + \mathcal{L}_{\text{kl}}$ (Baseline) & 19.93 & 26.56 & 15.96 & 0.841 & 0.210 \\
E002 & + $\mathcal{L}_{\text{kl\_ref}}$ & 24.46 & \best{33.50} & 20.46 & \best{0.886} & \best{0.159} \\
E003 & Full loss (Eq.~\ref{eq:total}): + $\mathcal{L}_{\text{comp}}$ + $\mathcal{L}_{\text{hard\_alpha}}$ & \best{25.53} & 32.80 & \best{22.14} & 0.880 & 0.165 \\
\bottomrule
\end{tabular}
}
\end{table}

\textbf{Key findings:} (1) The baseline (E001) using only standard L1 + perceptual + KL losses establishes a weak starting point (19.93~dB PSNR), indicating that standard reconstruction objectives are insufficient for high-fidelity alpha learning. (2) Adding $\mathcal{L}_{\text{kl\_ref}}$ (E002) provides the single largest improvement (+4.53~dB PSNR) by anchoring the latent space to the pretrained model, maintaining structural consistency. (3) Adding the composite loss $\mathcal{L}_{\text{comp}}$ and $\mathcal{L}_{\text{hard\_alpha}}$ (E003) further improves alpha quality (+1.68~dB PSNR\textsubscript{alpha}) by teaching the model to understand transparency through random background compositing.

Despite its extremely small weight ($10^{-10}$), $\mathcal{L}_{\text{kl\_ref}}$ plays an outsized role in representation learning. This demonstrates that even a minimal soft anchor to the original latent space provides sufficient regularization to prevent catastrophic forgetting---a simple yet highly effective mechanism compared to the bidirectional distribution shifting required by existing methods.
Additional ablation studies regarding the layer unfreezing strategy, training resolution, and frame count are provided in Appendix~\ref{sec:appendix_ablations}.

\subsection{DiT Fine-tuning with Standard DoRA}
\label{sec:lora_dora}

To evaluate downstream generation after adapting Wan2.1 for RGBA, we fine-tune the DiT with standard DoRA adapters (see Section~\ref{sec:dit}) on approximately 2K RGBA video-caption pairs (2,357 samples). Captions are generated by Qwen-VL and verified by human annotators, and we hold out 100 RGBA videos for testing.
We report automatic metrics from our internal evaluation suite, including \emph{Aesthetic}, \emph{Text Alignment}, \emph{Naturalness}, \emph{Motion Smoothness}, and \emph{Temporal Flickering}; for RGB generation we additionally report \emph{Dynamic Degree} and \emph{Temporal Spatial Consistency}. Table~\ref{tab:lora_dora} summarizes T2V comparisons.

\begin{table}[t]
\centering
\caption{\textbf{Generation results on RGBA and RGB evaluation sets.} \best{Best} results are shown in bold. For RGBA T2V, we compare our DoRA variant against Wan-Alpha and TransPixeler. For RGB T2V, we compare our pipeline against the original Wan2.1 to assess the effect of VAE replacement on standard RGB generation.}
\label{tab:lora_dora}
\vspace{0.5em}
\resizebox{\textwidth}{!}{
\begin{tabular}{l c c c c c}
\toprule
\textbf{Metric} & \multicolumn{3}{c}{\textbf{RGBA, T2V}} & \multicolumn{2}{c}{\textbf{RGB, T2V}} \\
\cmidrule(lr){2-4}\cmidrule(lr){5-6}
& \textbf{Wan-Alpha} & \textbf{TransPixeler} & \textbf{Ours} & \textbf{Ours} & \textbf{Wan2.1} \\
\midrule
\textbf{Aesthetic Quality}$\uparrow$ & \best{0.5232} & 0.4408 & 0.4694 & \best{0.6116} & 0.5471 \\
\textbf{Text Alignment}$\uparrow$ & 3.9794 & 3.2188 & \best{4.2600} & 3.8889 & \best{4.0566} \\
\textbf{Naturalness}$\uparrow$ & 3.6939 & 3.0156 & \best{4.0200} & \best{3.9821} & 3.8438 \\
\textbf{Motion Smoothness}$\uparrow$ & 0.9927 & 0.9936 & \best{0.9948} & 0.9896 & \best{0.9898} \\
\textbf{Temporal Flickering}$\uparrow$ & 0.9878 & 0.9905 & \best{0.9928} & \best{0.9823} & 0.9800 \\
\textbf{Dynamic Degree (RGB)}$\uparrow$ & -- & -- & -- & 0.3846 & \best{0.4800} \\
\textbf{Temporal Spatial Consistency (RGB)}$\uparrow$ & -- & -- & -- & \best{0.2388} & 0.2283 \\
\bottomrule
\end{tabular}
}
\end{table}

Table~\ref{tab:lora_dora} reports the current generation results for RGBA T2V and RGB T2V. On RGBA T2V, our DoRA variant significantly outperforms Wan-Alpha on most metrics: we achieve superior performance on \best{Text Alignment} (\best{4.2600} vs.\ 3.9794), \best{Naturalness} (\best{4.0200} vs.\ 3.6939), \best{Motion Smoothness} (\best{0.9948} vs.\ 0.9927), and \best{Temporal Flickering} (\best{0.9928} vs.\ 0.9878), while remaining competitive on Aesthetic Quality (0.4694 vs.\ 0.5232).
Crucially, on RGB T2V, we evaluate our system in a \emph{zero-shot} setting: we simply replace the original Wan2.1 VAE with our UniRGBA-VAE without performing any further fine-tuning on the DiT for RGB data. As shown in the table, our pipeline preserves---and on several metrics even surpasses---the original pretrained Wan2.1 backbone: our variant scores higher on \best{Aesthetic Quality} (0.6116 vs.\ 0.5471), \best{Naturalness} (3.9821 vs.\ 3.8438), \best{Temporal Flickering} (0.9823 vs.\ 0.9800), and \best{Temporal Spatial Consistency} (0.2388 vs.\ 0.2283), while remaining closely matched on Motion Smoothness (0.9896 vs.\ 0.9898). These results confirm that our UniRGBA-VAE acts as a true drop-in replacement that strictly preserves the original RGB latent space, allowing the pretrained DiT to generate high-quality RGB videos out-of-the-box.

\section{Qualitative Results}
\label{sec:qualitative}

\begin{figure}[htbp]
\centering
\sbox{0}{\includegraphics{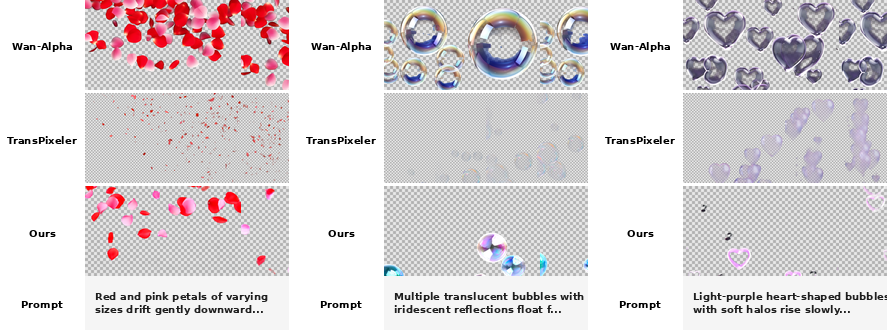}}
\edef\promptcrop{\the\dimexpr\ht0*16/100\relax}
\includegraphics[width=0.90\linewidth,trim=0 {\promptcrop} 0 0,clip]{figures/qualitative/t2v_three_way_compare.png}\par\smallskip
\begin{minipage}{0.90\linewidth}
\raggedright\small
\textbf{Left:} Red and pink petals of varying sizes drift gently downward\ldots\par
\textbf{Center:} Multiple translucent bubbles with iridescent reflections float f\ldots\par
\textbf{Right:} Light-purple heart-shaped bubbles with soft halos rise slowly\ldots
\end{minipage}
\caption{
\textbf{Three-way qualitative comparison on visual-effect T2V prompts.}
Wan-Alpha (top), TransPixeler (middle), and Ours (bottom, DoRA on Wan2.1-T2V-14B) are stacked, all rendered on a checkerboard background to visualize transparency. Our method produces transparency and color content that is visually competitive with Wan-Alpha while clearly outperforming TransPixeler on challenging non-elliptical visual-effect scenes.
}
\label{fig:qual_t2v}
\end{figure}

Figure~\ref{fig:qual_t2v} shows representative three-way comparisons on three visual-effect T2V prompts, arranged in a single row. For each case, we vertically stack the results from \textbf{Wan-Alpha} (full fine-tuning of Wan2.1-T2V-14B), \textbf{TransPixeler}, and \textbf{Ours} (DoRA adapters on Wan2.1-T2V-14B). Each row shows 4 uniformly sampled frames composited onto a checkerboard background so that semi-transparent regions remain visible. The prompt associated with each case is listed beneath its block. To complement these generation examples, we also provide additional selected qualitative reconstruction demos in Appendix Figures~\ref{fig:appendix_recon_demo_1}--\ref{fig:appendix_recon_demo_3}, as well as a supplementary qualitative I2V visualization in Appendix Figure~\ref{fig:appendix_i2v_demo}. Each appendix reconstruction demo is shown as a triptych containing black-background compositing, checkerboard compositing, and the predicted alpha matte, making RGB fidelity and transparency quality easy to inspect jointly.

\section{Conclusion}
\label{sec:conclusion}

We present \textbf{UniRGBA}, a framework that adds native RGBA support to pretrained video diffusion models through minimal intervention. By treating alpha as a fourth channel and anchoring the fine-tuned VAE posterior to the frozen RGB prior via a KL reference loss, we preserve the pretrained latent manifold by construction---avoiding the catastrophic forgetting that plagues methods relying on latent restructuring or dedicated alpha streams. \textbf{UniRGBA-VAE} achieves state-of-the-art RGBA reconstruction, and lightweight \textbf{DoRA adaptation} of the DiT is sufficient for downstream generation: our pipeline outperforms current baselines on RGBA T2V, extends naturally to I2V, and acts as a zero-shot drop-in replacement with no degradation in RGB quality. Limitations and future directions are discussed in Appendix~\ref{sec:appendix_limitation}.

\bibliographystyle{plain}
\bibliography{references}

\newpage
\appendix
\renewcommand{\theHfigure}{appendix.\arabic{figure}}
\renewcommand{\theHtable}{appendix.\arabic{table}}
\setcounter{figure}{0}
\setcounter{table}{0}
\renewcommand{\thefigure}{S\arabic{figure}}
\renewcommand{\thetable}{S\arabic{table}}

\section*{\centering{Appendix}} \label{appendix}

This Appendix is organized into the following sections: Section~\ref{sec:appendix_setup} provides the detailed experimental setup, including dataset construction and training schedules. Section~\ref{sec:appendix_ablations} presents additional ablation studies on the layer unfreezing strategy as well as the training resolution and frame count. Section~\ref{sec:appendix_visual_demos} shows additional qualitative reconstruction demos. Section~\ref{sec:appendix_limitation} discusses the limitations and future work of UniRGBA.

\section{Detailed Experimental Setup}
\label{sec:appendix_setup}

\subsection{Datasets}
\label{sec:appendix_datasets}

We curate an internal RGBA video dataset from historical special-effect asset packages used in live-streaming production, covering a wide range of \emph{visual-effect} transparent content (e.g., particles, flame, smoke, fireworks, animated stickers, and virtual gifts). We construct RGBA samples by extracting and standardizing alpha-bearing assets, then perform a two-stage labeling pipeline: (1) automatic captioning with Qwen-VL~\cite{team2023qwen}, followed by (2) manual verification and filtering to remove low-quality, ambiguous, or duplicated samples.

We use different subsets for reconstruction and generation experiments:
\begin{itemize}[leftmargin=2em]
\item \textbf{VAE ablation set}: 1K RGBA videos for rapid loss ablations and other small-scale experimental iterations.
\item \textbf{VAE main training set}: a curated high-quality RGBA subset with 17K videos, used for the final reconstruction model reported in Table~\ref{tab:main}.
\item \textbf{T2V generation set}: approximately 2K RGBA videos (2,357 samples) with Qwen-VL captions verified by human annotators for DoRA training, plus a held-out test set of 100 RGBA videos for generation evaluation.
\item \textbf{WebVid evaluation sets}: for RGB-only VAE reconstruction, we use a 5K-sample WebVid test set; for RGB T2V generation, we use a manually selected 100-sample WebVid test subset.
\end{itemize}

\subsection{Implementation Details}
\label{sec:appendix_impl}

\textbf{VAE Training.} All VAE experiments use the AdamW optimizer on 32 GPUs with a per-GPU batch size of 1. Because our dataset contains mixed resolutions, we normalize each sample by resizing while preserving aspect ratio and applying center-crop/pad to the target training resolution. Unless otherwise specified, we use 5 frames at 832$\times$480 for the loss ablation, 128$\times$128 for the resolution and layer-unfreezing ablations on the 1K ablation set, and scale to 832$\times$480 for the final reconstruction model trained on the 17K high-quality subset.

\textbf{Two-Stage Schedule.} For the final reconstruction model, we train for 100 epochs on the 17K high-quality subset under a single cosine learning-rate schedule with 640 warmup steps, peaking at $5\times10^{-5}$ after warmup and decaying to $1.5\times10^{-5}$ at epoch 100. Training is split into two stages based on the loss configuration: \emph{Stage-1 (epochs 0--30)} uses only the six-term progressive reconstruction loss (Eq.~\ref{eq:total}); \emph{Stage-2 (epochs 30--100)} continues along the same cosine schedule (lr $\approx 4.3\times10^{-5}$ at the epoch-30 transition, decaying to $1.5\times10^{-5}$ at epoch 100) and additionally introduces a $0.02$-weighted GAN objective to refine fine-grained details. Unless otherwise specified, reported Stage-1/Stage-2 numbers correspond to the end-of-stage checkpoints of this schedule.

\section{Additional Ablation Studies}
\label{sec:appendix_ablations}

\subsection{Layer Unfreezing Strategy}
\label{sec:appendix_layer}

Table~\ref{tab:layer_ablation} compares different unfreezing strategies on an early data-scaling run. While the 2+1 configuration achieves the highest pixel-level PSNR, the symmetric 2+2 configuration obtains substantially better perceptual quality (LPIPS 0.082 vs.\ 0.0995) and structural similarity (SSIM 0.936 vs.\ 0.898). This indicates that unfreezing both the encoder and the decoder symmetrically allows the model to learn more coherent RGBA representations. We therefore select 2+2 as our default configuration.

\begin{table}[!htbp]
\centering
\caption{Layer unfreezing ablation. The $k_e$+$k_d$ notation denotes $k_e$ unfrozen encoder layers and $k_d$ unfrozen decoder layers.}
\label{tab:layer_ablation}
\begin{tabular}{cl c c c c}
\toprule
\textbf{ID} & \textbf{Strategy ($k_e$+$k_d$)} & \textbf{PSNR}$\uparrow$ & \textbf{SSIM}$\uparrow$ & \textbf{LPIPS}$\downarrow$ & \textbf{MSE}$\downarrow$ \\
\midrule
E004 & 1+1          & 28.12          & 0.896          & 0.112           & 0.0076 \\
E005 & 2+1          & \textbf{29.04} & 0.898          & 0.0995          & \textbf{0.0023} \\
E006 & 1+2          & 27.50          & 0.910          & 0.111           & 0.0085 \\
E007 & \textbf{2+2} & 26.63          & \textbf{0.936} & \textbf{0.082}  & 0.0017 \\
\bottomrule
\end{tabular}
\end{table}

\subsection{Training Resolution and Frame Count}
\label{sec:appendix_res_frame}

Table~\ref{tab:res_frame_ablation} isolates the effects of training resolution and frame count. Fixing the temporal window to 5 frames reveals an inverse relationship between spatial resolution and reconstruction metrics: lower resolutions generally yield higher PSNR and lower LPIPS. We hypothesize this occurs because a fixed-capacity VAE can focus more effectively on learning the underlying alpha channel structure when relieved of the burden of modeling high-frequency spatial details. Fixing the resolution to $256\times256$ shows a strict monotonic decay in reconstruction quality as the sequence length grows from 5 to 21 frames, indicating that shorter temporal windows allow the model to achieve better per-frame compression fidelity.

\begin{table}[!htbp]
\centering
\caption{Resolution and frame count ablations. \textbf{Left:} effect of spatial resolution with 5 frames. \textbf{Right:} effect of temporal length at $256\times256$.}
\label{tab:res_frame_ablation}
\begin{minipage}[t]{0.48\textwidth}
\centering
\small
\setlength{\tabcolsep}{3pt}
\begin{tabular}[t]{c c c c c}
\toprule
\textbf{Resolution} & \textbf{PSNR}$\uparrow$ & \textbf{SSIM}$\uparrow$ & \textbf{LPIPS}$\downarrow$ & \textbf{ID} \\
\midrule
832$\times$480 & 25.53          & 0.880          & 0.165          & E003 \\
256$\times$256 & \textbf{27.81} & 0.897          & 0.141          & E008 \\
128$\times$128 & 27.74          & \textbf{0.899} & \textbf{0.111} & E009 \\
\bottomrule
\end{tabular}
\end{minipage}\hfill
\begin{minipage}[t]{0.48\textwidth}
\centering
\small
\setlength{\tabcolsep}{3pt}
\begin{tabular}[t]{c c c c c}
\toprule
\textbf{Frames} & \textbf{PSNR}$\uparrow$ & \textbf{SSIM}$\uparrow$ & \textbf{LPIPS}$\downarrow$ & \textbf{ID} \\
\midrule
5  & \textbf{27.81} & \textbf{0.897} & \textbf{0.141} & E008 \\
9  & 26.36          & 0.890          & 0.157          & E010 \\
13 & 25.91          & 0.886          & 0.162          & E011 \\
21 & 24.65          & 0.883          & 0.167          & E012 \\
\bottomrule
\end{tabular}
\end{minipage}
\end{table}

\clearpage
\section{Additional Qualitative Reconstruction Demos}
\label{sec:appendix_visual_demos}

\begin{figure}[!htbp]
\centering
\includegraphics[width=0.95\linewidth]{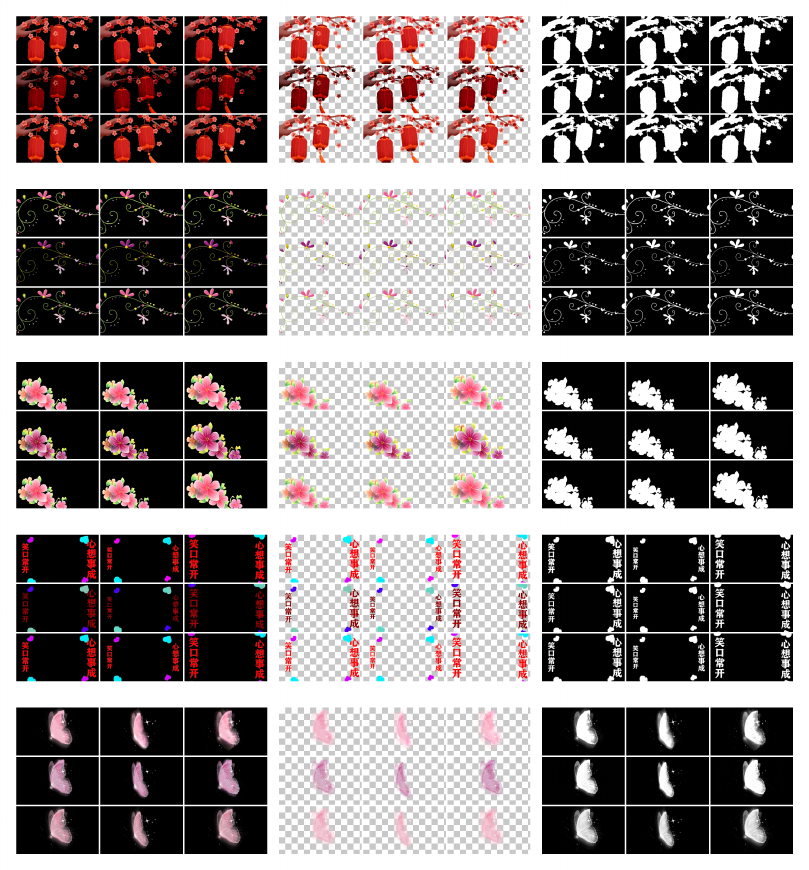}
\caption{
\textbf{Additional qualitative reconstruction demos (page 1/3).}
Five selected held-out RGBA videos. Within each triptych, the same example is shown as RGB composited on a black background (left), RGB composited on a checkerboard background (center), and the alpha matte (right). Within each panel, rows correspond to UniRGBA reconstruction (top), Wan-Alpha reconstruction (middle), and the ground-truth input video (bottom). The examples highlight strong color preservation together with clean and spatially coherent transparency on diverse visual-effect assets.
}
\label{fig:appendix_recon_demo_1}
\end{figure}

\begin{figure}[p]
\centering
\includegraphics[width=\linewidth]{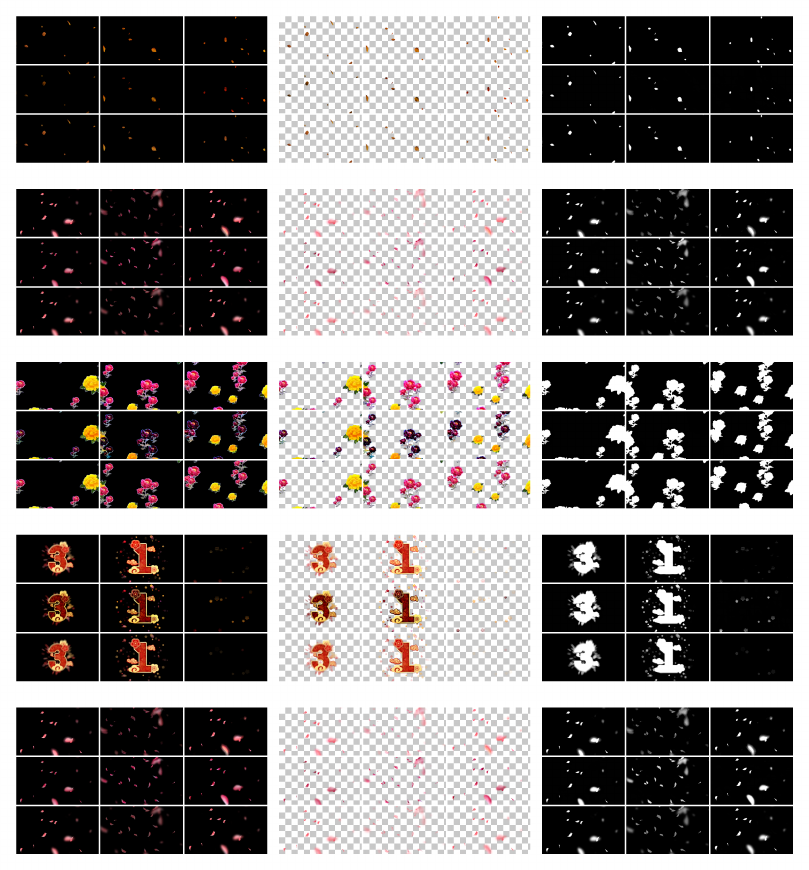}
\caption{
\textbf{Additional qualitative reconstruction demos (page 2/3).}
Five more selected held-out RGBA videos, shown in the same triptych format as Figure~\ref{fig:appendix_recon_demo_1}. Rows correspond to UniRGBA reconstruction, Wan-Alpha reconstruction, and ground truth. Across these cases, UniRGBA preserves both high-frequency color structure and soft alpha transitions that are critical for live-streaming visual-effect assets.
}
\label{fig:appendix_recon_demo_2}
\end{figure}

\begin{figure}[p]
\centering
\includegraphics[width=\linewidth]{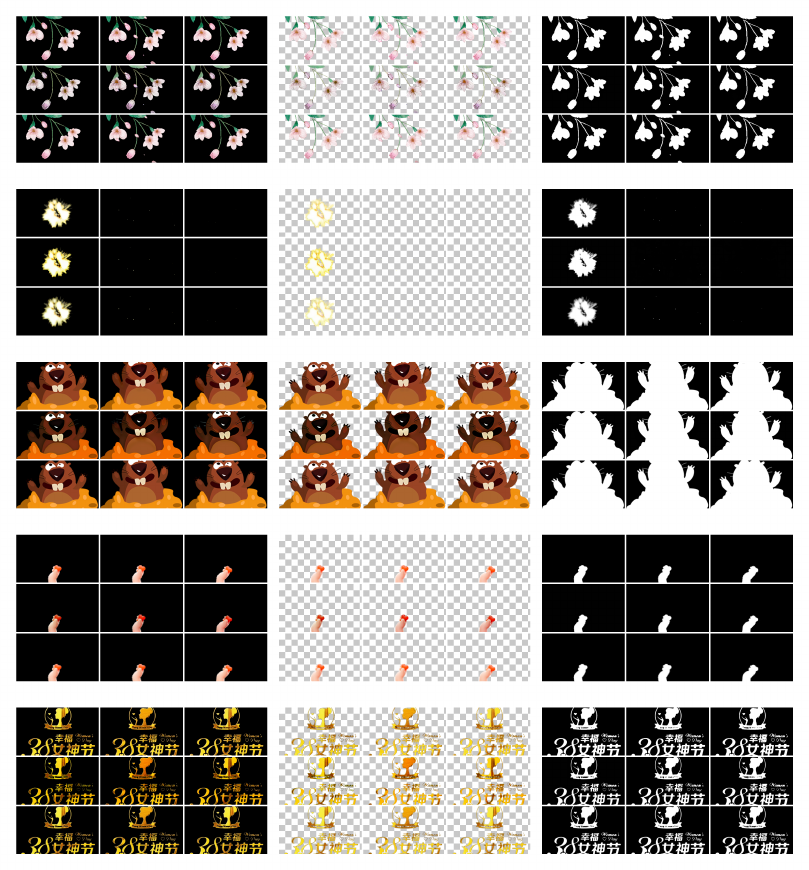}
\caption{
\textbf{Additional qualitative reconstruction demos (page 3/3).}
Five further selected held-out RGBA videos, again shown as black-composited RGB, checkerboard-composited RGB, and alpha, with rows corresponding to UniRGBA reconstruction, Wan-Alpha reconstruction, and ground truth. These additional examples reinforce the robustness of UniRGBA on challenging semi-transparent structures, thin boundaries, and complex visual-effect textures.
}
\label{fig:appendix_recon_demo_3}
\end{figure}

As a complement to the quantitative reconstruction results in Section~\ref{sec:main_results}, Figures~\ref{fig:appendix_recon_demo_1}--\ref{fig:appendix_recon_demo_3} present 15 selected held-out RGBA examples. For each example, we show a triptych consisting of RGB composited on a black background, RGB composited on a checkerboard background, and the alpha matte. Within each panel, the first row shows the reconstruction produced by \textbf{UniRGBA}, the second row shows the corresponding reconstruction from \textbf{Wan-Alpha}, and the third row shows the \textbf{ground-truth input video}. This layout makes it easy to jointly inspect foreground appearance, semi-transparent boundaries, and the consistency between RGB appearance and alpha structure across methods.

\clearpage
\section{Supplementary I2V Visualization}
\label{sec:appendix_i2v}

\begin{figure}[!htbp]
\centering
\includegraphics[width=\linewidth]{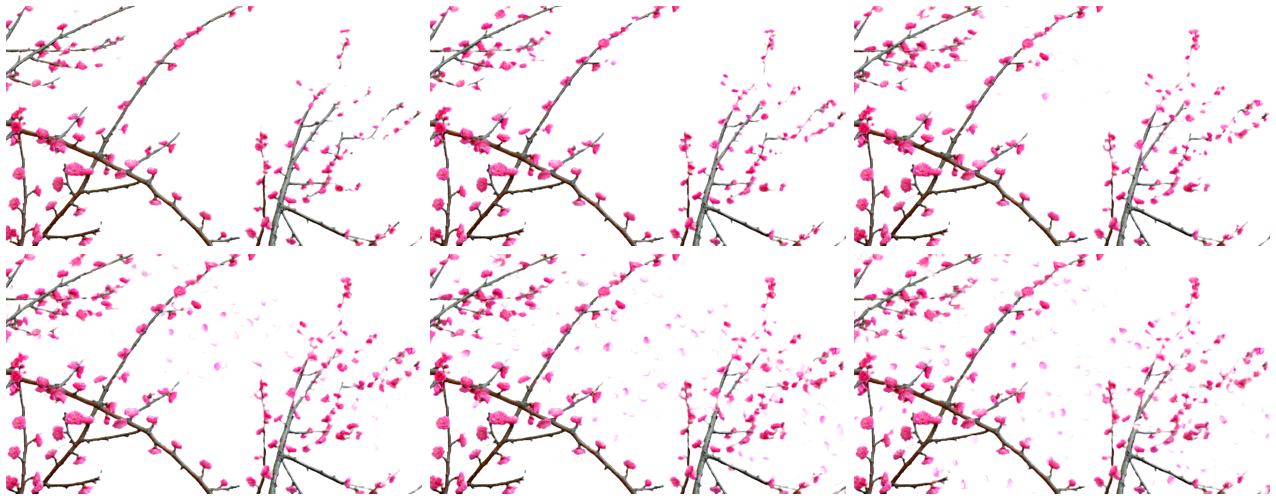}
\caption{
\textbf{Supplementary qualitative RGBA I2V result.}
Six uniformly sampled frames from an RGBA image-to-video generation example. The prompt is: ``A tree branch is covered with pink blossoms, and petals slowly drift through the air. The branch is dark brown, the flowers are densely distributed, and some petals have already fallen, forming a trajectory extending from the lower-left to the upper-right. The petals are pink and white, swaying gently in the wind.'' The generated sequence maintains coherent blossom appearance while producing temporally smooth petal motion and soft semi-transparent details.
}
\label{fig:appendix_i2v_demo}
\end{figure}

Figure~\ref{fig:appendix_i2v_demo} provides an additional qualitative RGBA I2V example. We show 6 uniformly sampled frames from the generated video to illustrate temporal coherence, alpha-aware motion evolution, and the preservation of delicate semi-transparent structures over time.

\section{Limitation and Future Work}
\label{sec:appendix_limitation}

Despite the strong performance demonstrated by UniRGBA, several limitations remain and open avenues for future research. First, the current RGBA training set is limited in scale ($\sim$2K videos for DoRA training) and is focused on visual-effect content; expanding coverage to more diverse object- and scene-centric RGBA data could further strengthen both reconstruction and generation quality. Second, our current pipeline is evaluated at 832$\times$480; scaling the unified VAE and DiT adapter to higher spatial resolutions (e.g., 720p/1080p) and longer temporal windows remains an open engineering direction. Finally, we restrict adaptation to lightweight DoRA for controlled comparison with Wan-Alpha; exploring full-parameter fine-tuning and larger adapter capacity may further close the gap on high-level aesthetic attributes observed in full-finetune baselines.

\end{document}